\documentclass[journal]{IEEEtran}
\newif\ifarxiv
\arxivtrue

\usepackage[T1]{fontenc}
\usepackage{graphicx}
\usepackage{booktabs}
\usepackage{tabularx}
\usepackage{adjustbox}
\usepackage{amsmath}
\usepackage{amssymb}
\usepackage{amsfonts}
\usepackage{mathtools}
\usepackage{microtype}
\usepackage{xcolor}
\usepackage{colortbl}
\usepackage{xspace}
\usepackage{etoolbox}
\usepackage{multicol}
\raggedcolumns
\usepackage{pgfplots}\pgfplotsset{compat=1.18}
\usepackage{tikz}
\usetikzlibrary{arrows.meta,positioning,shapes.geometric,calc,fit,backgrounds}

\usepackage[numbers,sort&compress]{natbib}

\usepackage{hyperref}
\hypersetup{hidelinks, breaklinks=true,
  pdftitle={Know the Design, Pick the Attack: Sentence-Keyed Watermarks for LLMs},
  pdfauthor={Jon Carmona, Jon Irureta, Jon Imaz, Aizea Lojo, Juan Martinez-Romo, Lourdes Araujo}}
\usepackage[capitalise, nameinlink]{cleveref}
\crefname{appendix}{Appendix}{Appendices}\Crefname{appendix}{Appendix}{Appendices}

\input{macros}

\begin{document}

\title{Know the Design, Pick the Attack:\\ Sentence-Keyed Watermarks for LLMs}

\author{Jon~Carmona, Jon~Irureta, Jon~Imaz, Aizea~Lojo, Juan~Mart\'inez-Romo, and~Lourdes~Araujo%
\ifarxiv\thanks{This work has been submitted to the IEEE for possible publication. Copyright may be
transferred without notice, after which this version may no longer be accessible.}\fi%
\thanks{This work has been partially supported by the Spanish Ministry of Science and Innovation
within the EDHER-MED Project under grant PID2022-136522OB-C21, as well as within the MIND-SIGHT
Project under Grant PID2025-167847OB-C21.}%
\thanks{J.~Carmona, J.~Irureta, J.~Imaz and A.~Lojo are with IKERLAN Technology Research Centre,
Basque Research and Technology Alliance (BRTA), Arrasate-Mondrag\'on, Spain.}%
\thanks{J.~Carmona, J.~Mart\'inez-Romo and L.~Araujo are with Universidad Nacional de Educaci\'on a
Distancia (UNED), Madrid, Spain.}%
\ifarxiv\thanks{Corresponding author: Jon Carmona.}\else
\fi}

\ifarxiv
\markboth{}{}
\else
\markboth{IEEE Transactions on Information Forensics and Security}%
{Carmona \MakeLowercase{\textit{et al.}}: Sentence-Keyed Watermarks for LLMs}
\fi

\maketitle

\begin{abstract}
A watermark lets the provider of a language model recognise the text its model wrote, which
matters more as generated text becomes hard to tell from human writing: provenance underpins
academic integrity, the fight against misinformation and new regulation. Providers have started to
watermark the text of their models, and to explain how their watermarks work. Explaining a
watermark, however, also tells an attacker how to break it, and there are two ways to do so:
reword the text to erase the mark, or learn the mark and copy it onto text the model never wrote.
Each attack has been studied on its own. We study an attacker that reads the published design to
choose between them: it erases the watermarks that are hard to learn and forges the ones that
survive rewording. Against it, every existing watermark we compare fails. We propose \KL, which reads its secret
key from the grammar of the previous sentence, which rewording changes little, and whose detector
does not need to recover the key exactly. It outperforms every watermark we compare on both models
we test, losing about half of its signal where the best alternative loses two thirds, at no extra cost
in text quality over the best-known watermarks.

\end{abstract}


\section{Introduction}\label{sec:intro}

\IEEEPARstart{L}{anguage} models now produce text that is hard to distinguish from human writing, which makes
establishing its provenance increasingly important. Provenance underpins academic integrity and the
fight against misinformation, it is a legal requirement under the EU AI Act \citep{euaiact2024},
and it lets developers keep generated text out of training data, where it degrades the models
trained on it \citep{shumailov2024collapse}. Establishing it after the fact, however, is
unreliable: classifiers built to separate human from generated text fail under simple paraphrasing
\citep{sadasivan2023}. A watermark avoids the problem by establishing provenance at the source,
since the model leaves an invisible signal in the text as it writes. Google, Anthropic and OpenAI
now watermark the text their models generate in this way
\citep{dathathri2024synthid,anthropic2026watermark,li2026textgrain}.

\begin{figure}[t]
  \centering
  \begin{minipage}[b]{0.64\linewidth}
    \centering
    \definecolor{vizKL}{HTML}{2A78D6}
\definecolor{vizBase}{HTML}{8F8D87}
\definecolor{vizOut}{HTML}{D3D1CB}
\begin{tikzpicture}
\begin{axis}[
  width=\linewidth, height=4.4cm, scale only axis=false,
  xmin=0, xmax=102, ymin=0, ymax=100,
  xtick={0,50,100}, xticklabels={0,50,100},
  ytick={0,50,100}, yticklabels={0,50,100},
  axis line style={black!40},
  tick label style={font=\scriptsize, color=black!80},
  label style={font=\scriptsize},
  xlabel={Forged (\%)}, ylabel={Erased by paraphrase (\%)},
  xlabel style={yshift=3pt}, ylabel style={yshift=-4pt},
  xmajorgrids, ymajorgrids, major grid style={black!7},
  clip=false,
]
\addplot[black!45, line width=0.7pt, mark=none]
  coordinates {(4.8,57.1) (29.8,38.5) (75.5,24.5) (80.7,18.4) (96.8,8.1)};
\addplot[only marks, mark=*, mark size=1.8pt, mark options={fill=vizOut, draw=black!30}]
  coordinates {(4.7,79.3) (4.8,79.1) (95.5,45.9) (97.0,45.7)};
\addplot[only marks, mark=*, mark size=1.8pt, mark options={fill=vizBase, draw=black!60}]
  coordinates {(4.8,57.1) (75.5,24.5) (80.7,18.4) (96.8,8.1)};
\addplot[only marks, mark=*, mark size=2.6pt, mark options={fill=vizKL, draw=vizKL}]
  coordinates {(29.8,38.5)};
\addplot[draw=none, mark=none, nodes near coords, point meta=explicit symbolic,
  every node near coord/.append style={anchor=west, xshift=1pt, font=\tiny, black!85}]
  coordinates {(4.8,57.1) [SynthID]};
\addplot[draw=none, mark=none, nodes near coords, point meta=explicit symbolic,
  every node near coord/.append style={anchor=south east, xshift=0pt, yshift=1pt, font=\tiny, black!85}]
  coordinates {(75.5,24.5) [SIR]};
\addplot[draw=none, mark=none, nodes near coords, point meta=explicit symbolic,
  every node near coord/.append style={anchor=south west, xshift=0pt, yshift=1pt, font=\tiny, black!85}]
  coordinates {(80.7,18.4) [KGW-1]};
\addplot[draw=none, mark=none, nodes near coords, point meta=explicit symbolic,
  every node near coord/.append style={anchor=east, xshift=-2.5pt, yshift=-1pt, font=\tiny, black!85}]
  coordinates {(96.8,8.1) [Unigram]};
\addplot[draw=none, mark=none, nodes near coords, point meta=explicit symbolic,
  every node near coord/.append style={anchor=west, xshift=1pt, font=\tiny, black!50}]
  coordinates {(4.7,79.3) [DiPmark, KGW-4]};
\addplot[draw=none, mark=none, nodes near coords, point meta=explicit symbolic,
  every node near coord/.append style={anchor=south east, xshift=2pt, yshift=2pt, font=\tiny, black!50}]
  coordinates {(96.2,45.8) [(k-)SemStamp]};
\addplot[draw=none, mark=none, nodes near coords, point meta=explicit symbolic,
  every node near coord/.append style={anchor=south west, xshift=1pt, yshift=1pt,
    font=\tiny\bfseries, text=vizKL}]
  coordinates {(29.8,38.5) [KeyedLex]};
\end{axis}
\end{tikzpicture}\\[-2pt]
    {\scriptsize (a)}
  \end{minipage}\hfill
  \begin{minipage}[b]{0.34\linewidth}
    \centering
    \definecolor{vizKL}{HTML}{2A78D6}
\definecolor{vizBase}{HTML}{8F8D87}
\begin{tikzpicture}
\begin{axis}[
  xbar, bar width=6.2pt, scale only axis, width=0.80\linewidth, height=3.13cm,
  xmin=40, xmax=100, enlarge y limits=0.06,
  xtick={40,70,100}, xticklabels={40,70,100},
  symbolic y coords={k-SemStamp,SemStamp,Unigram,KGW-1,KGW-4,DiPmark,SIR,SynthID,KeyedLex},
  ytick=\empty,
  axis line style={black!40}, axis x line*=bottom, axis y line=none,
  xticklabel style={font=\scriptsize}, label style={font=\scriptsize},
  xlabel={Weaker axis (\%)}, xlabel style={yshift=3pt},
  xmajorgrids, major grid style={black!7},
  clip=true,
]
\addplot[fill=vizKL!16, draw=none, bar shift=0pt] coordinates {
  (100,k-SemStamp) (100,SemStamp) (93,KGW-4) (91,DiPmark) (66,SynthID)};
\addplot[fill=black!13, draw=none, bar shift=0pt] coordinates {
  (100,Unigram) (94,KGW-1) (87,SIR)};
\addplot[fill=vizKL!45, draw=none, bar shift=0pt] coordinates {(53,KeyedLex)};
\node[anchor=west, font=\tiny, black!85, inner sep=1pt] at (axis cs:40,SynthID) {SynthID};
\node[anchor=west, font=\tiny, black!85, inner sep=1pt] at (axis cs:40,SIR) {SIR};
\node[anchor=west, font=\tiny, black!85, inner sep=1pt] at (axis cs:40,DiPmark) {DiPmark};
\node[anchor=west, font=\tiny, black!85, inner sep=1pt] at (axis cs:40,KGW-4) {KGW-4};
\node[anchor=west, font=\tiny, black!85, inner sep=1pt] at (axis cs:40,KGW-1) {KGW-1};
\node[anchor=west, font=\tiny, black!85, inner sep=1pt] at (axis cs:40,Unigram) {Unigram};
\node[anchor=west, font=\tiny, black!85, inner sep=1pt] at (axis cs:40,SemStamp) {SemStamp};
\node[anchor=west, font=\tiny, black!85, inner sep=1pt] at (axis cs:40,k-SemStamp) {k-SemStamp};
\node[anchor=west, font=\tiny\bfseries, text=vizKL!55!black, inner sep=1pt] at (axis cs:40,KeyedLex) {KeyedLex};
\end{axis}
\end{tikzpicture}\\[-2pt]
    {\scriptsize (b)}
  \end{minipage}
  \caption{\textbf{\KL balances resistance to forgery and to paraphrase.} (a) Each point is a
  watermark, placed by how often it is forged and how much of its signal a paraphrase erases;
  lower left is better, and no watermark beats \KL on both axes. (b) The weaker axis of each
  watermark, in blue when erasure decides it and in grey when forgery does; lower is better, and
  the axis starts at 40\%. OPT-1.3B.}
  \label{fig:tradeoff}
\end{figure}
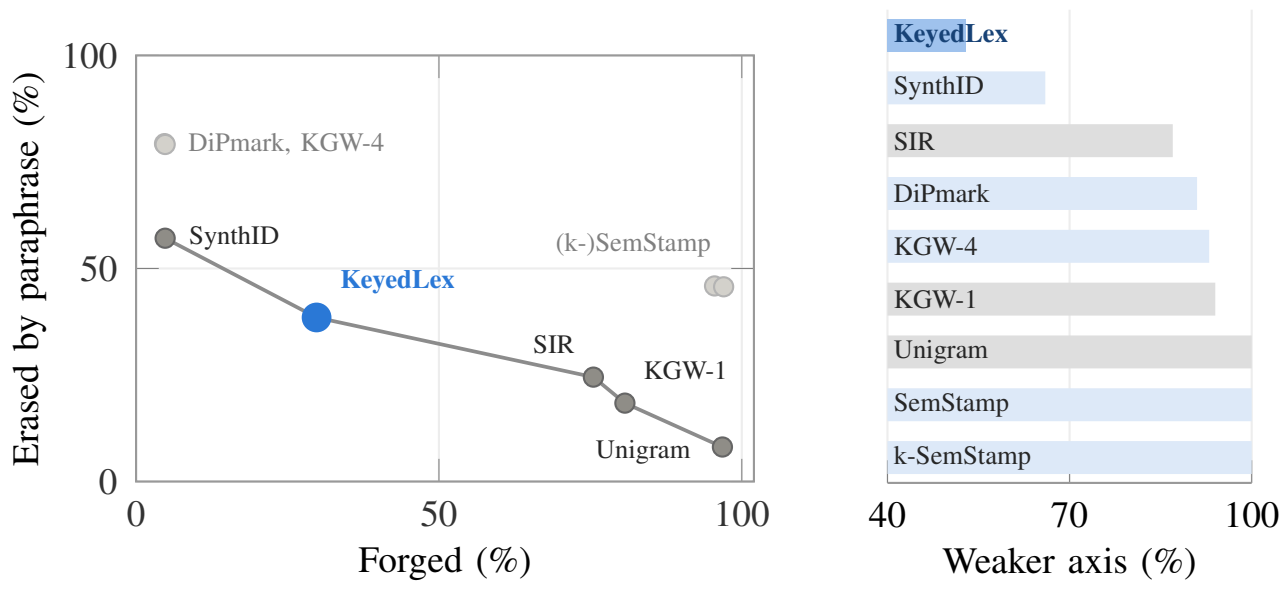

Most language models write text one token at a time. At each step several tokens could come next,
and the model picks one of them. A watermark changes that choice slightly. In the most widely
studied scheme \citep{kirchenbauer2023}, before each token is chosen, a secret key known only to
the provider divides all the tokens the model can use into two lists, and the model is nudged
towards one of them: the tokens in that list become a little more likely. Following the literature
\citep{kirchenbauer2023}, we call the list the model is nudged towards the \emph{green list}, and
the list of the remaining tokens the \emph{red list}. The nudge is small, and there is almost
always a suitable token in either list, so the text still reads naturally and a reader notices
nothing. Over a whole text, however, the nudge leaves a statistically significant trace, which
lets whoever holds the key tell whether a text was generated by the model. They count how many of
its tokens are green: in human writing that share is about what chance predicts, and in
watermarked text it is significantly higher. A statistical test decides whether the difference is
large enough, with a small, known chance of flagging human text by mistake.

The green list is one way of hiding a preference in the text, but not the only one. The literature
describes two ways of doing it \citep{liu2024survey}: changing how likely each token is (the
scheme we have just described), or changing how the next token is picked. The first
\citep{kirchenbauer2023,zhao2024,liu2024sir,wu2024dipmark} makes some tokens a little more
likely, and the model then picks as usual. The second kind leaves every token exactly as likely as
before, and uses the key only to decide which of the candidates is picked, in a way that looks
random to anyone without the key
\citep{kuditipudi2024robust,hou2024semstamp,dathathri2024synthid,li2026textgrain}. Both are
detected in the same way: whoever holds the key checks whether the text follows its preference more
often than chance would.

Whatever its kind, however, a watermark can be attacked in two ways. Someone who wants to pass generated text off as human
tries to \emph{erase} the mark, for instance by paraphrasing, which replaces green tokens with red
ones. Someone who wants to pin on the model a text it never wrote tries to \emph{forge} it: they
learn which tokens are green from watermarked text and write text of their own that the detector
accepts \citep{sadasivan2023,jovanovic2024}. Forgery is the more damaging of the two, because it
turns the watermark into false evidence against a provider or a person. Erasure and forgery are
thus two axes on which a watermark can fail, and a useful watermark has to resist both.

Both attacks have been studied. Erasure ranges from general paraphrasers \citep{krishna2023} to
paraphrasers tuned to evade watermarks
\citep{diaa2025adaptive,huang2025rlcracker,zhang2026character}; forgery learns the green list from
watermarked text \citep{sadasivan2023,jovanovic2024} or trains a model on it until the model
produces the mark \citep{gu2024learnability}. Each attack, however, is built for one of the two
axes, and none uses what is public about a watermark to decide which axis to attack, even though
much is public: Google, Anthropic and OpenAI have explained how their watermarks work
\citep{dathathri2024synthid,anthropic2026watermark,li2026textgrain}, and even a scheme nobody
discloses can be identified from a few queries \citep{gloaguen2025blackbox}. This is the setting
of Kerckhoffs's principle, a founding rule of security design
\citep{kerckhoffs1883,jovanovic2024}: a system must stay secure even when everything about it is
public except the key. For watermarks, its consequence has not been drawn: an attacker who knows how a watermark
is built knows which attack to use against it. If the key changes with every few tokens, an edit
breaks it, so the attacker paraphrases; if the key is stable, it learns it and forges. A watermark
can therefore look secure against the one attack nobody would use on it.

We propose a novel way to attack a watermark: read what its design discloses and attack the axis it
exposes. A watermark is then only as strong as its weaker axis, and we judge it by that axis
rather than by each axis apart. Judged this way, every existing watermark we compare fails on one of the two axes (\cref{fig:tradeoff}).
The family the providers deploy, which reads its key from the last few tokens, cannot be forged,
but paraphrase erases most of it. The watermarks that survive paraphrase are forged most of the
time.

We then propose \KL, a watermark built so that knowing its design helps the attacker little on
either axis. It chooses the green list from the grammar of the previous sentence and changes it
only once per sentence, which makes it hard to forge, and its detector tries every list instead of
recomputing the key, so rewording does not erase it. Against an attacker that adapts to each
design, \KL outperforms every watermark we compare, at no extra cost in text quality over the
best-known watermarks.

\paragraph{Contributions.}
\begin{itemize}
  \item \textbf{\KL}, a watermark whose green list is chosen by the grammar of the previous
        sentence, from a small set of secret lists, and whose detector tries each of them instead
        of recomputing the key. It is Pareto-optimal on forgery and erasure,
        and its weaker axis is the strongest of all the watermarks we compare, while raising
        perplexity less than KGW and Unigram.
  \item \textbf{A novel way to attack watermarks}, which exploits what each design discloses on both
        axes, and an evaluation that scores each watermark by its weaker axis, exposing
        vulnerabilities of every existing method we test.
  \item \textbf{An in-depth analysis of why \KL works}, which ties each of its properties to one
        design choice.
\end{itemize}
All code will be made publicly available upon acceptance.

\section{Background and Related Work}\label{sec:related}

\paragraph{Two kinds of watermark.} Watermarks for language models fall into two kinds
\citep{liu2024survey}: those that change how likely each token is, and those that
change how the next token is picked. In the first kind, at each step a key and the context select a
\emph{green list} $G$, a fraction $\gamma$ of the vocabulary, and the logits of its tokens get a
bias $\delta>0$ \citep{kirchenbauer2023,zhao2024,liu2024sir,wu2024dipmark}. In the second, the
probabilities stay as they are and the key decides which token is drawn
\citep{kuditipudi2024robust,hou2024semstamp,dathathri2024synthid,li2026textgrain}. Both are
detected the same way: measure how well a text of $T$ tokens follows the key and test whether it
does so more often than chance. For a green list, with $g$ the number of green tokens in the text,
the detector flags it when
\begin{equation}\label{eq:z}
  z=\frac{g-p_0 T}{\sqrt{T p_0(1-p_0)}} > z^{*},
\end{equation}
where $p_0\approx\gamma$ is the share of green tokens in human text and $z^{*}$ fixes the
false-positive rate.

\paragraph{Where the key comes from.} Watermarks differ mainly in what their key is read from and
how often it changes. Unigram \citep{zhao2024} fixes a single list. KGW \citep{kirchenbauer2023},
SynthID \citep{dathathri2024synthid} and DiPmark \citep{wu2024dipmark} derive it from a hash of
the preceding tokens and a secret, as do MorphMark \citep{wang2025morphmark},
which adapts the strength of KGW's bias, and EXP \citep{fernandez2023threebricks}. SIR \citep{liu2024sir}, its cross-lingual version X-SIR
\citep{he2024xsir} and Adaptive \citep{liu2024adaptive} derive it from an embedding of the
preceding text, and PostMark \citep{chang2024postmark} inserts words chosen from an embedding of the
whole text after generation. SemStamp \citep{hou2024semstamp} reads its key once per sentence: it
partitions a semantic embedding space and samples each sentence until its embedding falls in a
valid region, and k-SemStamp \citep{hou2024ksemstamp} replaces the random partition with $k$-means
clusters. \Cref{tab:families} summarises the watermarks we compare along these two properties.

\begin{table}[!htb]
  \centering
  \caption{Where each watermark reads its key from, and how often the key changes}
  \label{tab:families}
  \begin{tabular}{@{}lcc@{}}
  \toprule
  Watermark & Key read from & Changes \\
  \midrule
  Unigram \citep{zhao2024}             & None (fixed list)       & Never \\
  KGW-1 \citep{kirchenbauer2023}       & Previous token          & Every token \\
  KGW-4 \citep{kirchenbauer2023}       & Previous tokens         & Every token \\
  SynthID \citep{dathathri2024synthid} & Previous tokens         & Every token \\
  DiPmark \citep{wu2024dipmark}        & Previous tokens         & Every token \\
  SIR \citep{liu2024sir}               & Meaning of the prefix   & Every token \\
  SemStamp \citep{hou2024semstamp}     & Meaning of the sentence & Every sentence \\
  k-SemStamp \citep{hou2024ksemstamp}  & Meaning of the sentence & Every sentence \\
  \bottomrule
\end{tabular}

\end{table}

Watermarks can be attacked in two ways: an attacker either erases the mark or forges it.

\paragraph{Attacks that erase.} The simplest way to erase a watermark is to reword the text, so
that the tokens that carried the mark are replaced by others: paraphrase, notably with DIPPER
\citep{krishna2023}, a paraphraser trained to evade detectors of generated text, translation \citep{he2024xsir} and word-level edits such as synonym
substitution and typos \citep{kirchenbauer2023,piet2023markmywords}. Adaptive attacks tune the eraser against a known scheme. \citet{diaa2025adaptive} fine-tune a
paraphraser by preference optimisation to evade a given watermark and find that it also evades
others; RLCracker \citep{huang2025rlcracker} trains the paraphraser with reinforcement learning
from a few watermarked texts and no detector access; and \citet{zhang2026character} show that
character-level edits, guided by a reference detector, erase more for the same budget than token-
or sentence-level ones. All three adapt on the erasure axis only.

\paragraph{Attacks that forge.} \citet{sadasivan2023} showed that a watermark can be spoofed by
learning its green list from word frequencies in watermarked text. \citet{jovanovic2024} made this
a practical attack on the KGW family and Unigram: from watermarked text alone, an attacker learns
enough of the green lists both to forge the watermark and to erase it. \citet{gu2024learnability}
take a different route: they fine-tune a model on watermarked text until it produces the watermark
itself, and find that schemes keyed on short contexts are learnable while schemes keyed on long
ones are much harder to learn. \citet{pang2024} add piggyback spoofing, which inserts false or
harmful content into a genuine watermarked text without removing the mark. Each applies one fixed
recipe to every watermark; none tailors the forgery to how each watermark reads its key.

\paragraph{The trade-off.} What the key is read from decides which attack a watermark is exposed
to. A key read from a few tokens recurs often, so an attacker can learn it from watermarked text and
forge the watermark \citep{jovanovic2024,gu2024learnability}. A key read from many tokens is harder
to learn, but an edit to any of them changes it, so paraphrase erases the watermark
\citep{kirchenbauer2024reliability,shen2025seek}. Reading the key from meaning rather than words
aims to escape this, surviving paraphrase without becoming easy to learn
\citep{liu2024sir,hou2024semstamp}. Robustness and security therefore pull in opposite directions:
each design choice that helps one opens an attack on the other \citep{pang2024}, and it also moves
the quality of the text \citep{kirchenbauer2024reliability}. Designing a watermark is thus a
multi-objective problem, and a design has to be judged on both axes at once.

\paragraph{What is missing.} Erasure and forgery have so far been studied apart. The works that
consider both report each axis on its own \citep{jovanovic2024,pang2024,shen2025seek} or average
the two \citep{an2025contrastive}. Benchmarks that compare watermarks measure only erasure
\citep{piet2023markmywords,liang2025waterpark}, and so do the attacks that adapt to a watermark
\citep{diaa2025adaptive,huang2025rlcracker,zhang2026character}. No existing work attacks a
watermark on whichever axis its design leaves weaker, nor scores it by that axis. This paper closes
that gap, with an attack that exploits what each design discloses and with a watermark built to
withstand it.

\section{\KL: A Watermark Keyed on the Grammar of Each Sentence}\label{sec:methods}

\KL is built to resist an attacker who knows everything about the watermark except its secret
key. Knowing the design, the attacker can choose the attack it is weakest against, so \KL is
designed with three requirements in mind, one for each attack it must be robust against:
\begin{enumerate}
  \item \emph{Paraphrase.} If the key is read from the preceding tokens, the detector must
  recompute it from them, and an edit changes them \citep{kirchenbauer2024reliability}. The
  detector should not depend on recomputing the key exactly.
  \item \emph{Counting.} If the key changes with every token but repeats with its context, an
  attacker can count, for each context, which tokens are favoured \citep{jovanovic2024}. The key
  should change at a scale where observations cannot be aligned token by token.
  \item \emph{Averaging over keys.} If editing the text moves the key, an attacker can rewrite it
  under many keys and average the bias away \citep{pang2024}. Paraphrase should barely move the key.
\end{enumerate}

\begin{widetable}[tb]
  \centering
  \caption{Each watermark is attacked through its context}
  \label{tab:attack-families}
  \begin{tabular}{lccc}
  \toprule
  Watermark & Context & Forgery & Erasure \\
  \midrule
  Unigram \citep{zhao2024}             & None (fixed list)       & Count tokens       & Adaptive paraphrase \\
  KGW-1 \citep{kirchenbauer2023}       & Previous token          & Count per context  & Paraphrase \\
  KGW-4 \citep{kirchenbauer2023}       & Previous tokens         & Count per context  & Paraphrase \\
  SynthID \citep{dathathri2024synthid} & Previous tokens         & Count per context  & Paraphrase \\
  DiPmark \citep{wu2024dipmark}        & Previous tokens         & Count per context  & Paraphrase \\
  SIR \citep{liu2024sir}               & Meaning of the prefix   & Count per cluster  & Paraphrase \\
  SemStamp \citep{hou2024semstamp}     & Meaning of the sentence & Region transitions & Punctuation removal \\
  k-SemStamp \citep{hou2024ksemstamp}  & Meaning of the sentence & Region transitions & Punctuation removal \\
  \midrule
  \oursrow \KL (ours) & Previous sentence's grammar & Count per cell & Punctuation removal + Key averaging \\
  \bottomrule
\end{tabular}

\end{widetable}

\KL is a green-list watermark: generation adds a bias $\delta$ to the logits of the tokens in a
secret green list, a fraction $\gamma$ of the vocabulary, and the detector counts how many tokens
fall in it. What \KL changes is what selects the list. It reads the grammar of each sentence,
groups sentences of similar grammar, and gives each group its own list; a sentence is written with
the list of the previous sentence's group. The key thus changes once per sentence. It is read from
grammar rather than words because a paraphrase replaces the words of a sentence far more often than
it changes its grammar (\cref{app:grammar}). The detector tests each sentence against
every list and uses the grammar of the previous sentence only as a hint, so an edit that changes
that sentence does not by itself erase the mark.

\paragraph{The grammatical profile of a sentence.} \KL reads the grammar of a sentence $s$ from the
kinds of words it uses. To find them, a part-of-speech tagger labels each word with its word class,
and we keep the set $\mathcal{T}$ of the thirteen classes of Universal Dependencies
\citep{nivre2020ud} that carry grammar, leaving out punctuation, symbols, unknown words and
interjections. The \emph{profile} of $s$ is the share of its words in each class $t\in\mathcal{T}$:
\begin{equation}
  \mathbf{p}(s)=\big(p_t(s)\big)_{t\in\mathcal{T}}, \qquad \textstyle\sum_{t\in\mathcal{T}} p_t(s)=1 ,
\end{equation}
where $p_t(s)$ is the share of the words of $s$ in class $t$. In \texttt{``The cat sat on the mat''}, two of
the six words are determiners, two are nouns, one is a verb and one a preposition, so its profile
is $1/3$ each for determiners and nouns, $1/6$ each for verbs and prepositions, and $0$ for the
other nine classes. Two sentences with the same mix of word classes have the same profile, whatever
their words, so replacing words with synonyms, which keep their word class, leaves it unchanged unless the
tagger reads the new word differently.

\paragraph{From profile to green list.} Sentences with similar profiles share a green list. To fix
which profiles count as similar, \KL takes a sample of natural text, used in no evaluation, and
groups its sentences by their profiles with $k$-means. This is done once, before any text is
watermarked, and gives $k$ groups of sentences, which we call \emph{cells}. The choice of $k$ is discussed in
\cref{app:k}.
Profiles are compared after standardising each share against its spread in the sample. For each
of the thirteen classes $t$, the standardised share is
\begin{equation}
  \tilde{p}_t(s)=\frac{p_t(s)-\bar{p}_t}{\sigma_t},
\end{equation}
where $\bar{p}_t$ and $\sigma_t$ are the mean and standard deviation of $p_t$ over the sentences of
the sample, and the \emph{standardised profile} gathers them:
\begin{equation}
  \tilde{\mathbf{p}}(s)
  =\big(\tilde{p}_t(s)\big)_{t\in\mathcal{T}}.
\end{equation}
Without this
step the distance would depend almost only on the classes whose share changes most from one
sentence to the next, such as nouns, and would barely notice classes that are usually absent, such
as numbers. Each cell $j$ is summarised by its centre $\boldsymbol{\mu}_j$, the average of
$\tilde{\mathbf{p}}$ over its sentences, and any later sentence goes to the cell whose centre is
closest to its own standardised profile:
\begin{equation}
  c(s)=\arg\min_{j}\,\lVert \tilde{\mathbf{p}}(s)-\boldsymbol{\mu}_j\rVert .
\end{equation}
Each cell $j$ has its own green list
$G_j$, a pseudorandom fraction $\gamma$ of the vocabulary $V$ drawn from a secret seed,
independently of the other cells.

For example, \texttt{``The cat sat on the mat''} and \texttt{``The dog lay on the rug''} have the same profile and
fall in the same cell, so a sentence that follows either of them is written with the same list.
\texttt{``We think it will help''}, made mostly of pronouns and verbs, falls in another cell and selects
another list.

\paragraph{What is public and what is secret.} \KL assumes that everything about it is public except
one seed, as Kerckhoffs's principle asks of a cipher \citep{kerckhoffs1883}. The tagger, the
classes, the sample text and the cells are public: anyone can compute the profile of a sentence and
the cell it falls in, and so knows when the list changes. Only the content of the lists, drawn from
the seed, is secret. Grammar decides which list is active, not which tokens are in it, so an
attacker who wants to forge the mark has to learn the $k$ lists from watermarked text.

\paragraph{Generation.} While the model writes sentence $s_i$, let $j=c(s_{i-1})$ be the cell of the
previous sentence. The logit $\ell_v$ of each token $v$ becomes
\begin{equation}
  \ell'_v=\ell_v+\delta\,\mathbf{1}[v\in G_j],
\end{equation}
where $\mathbf{1}[\cdot]$ is the indicator function. When a sentence ends, the
tagger reads it and its cell selects the list for the next one. The first sentence has no
predecessor and is written without bias.

\paragraph{Detection.} The detector reads the text one sentence at a time. About a fraction
$\gamma$ of the tokens of a human sentence fall in any given list by chance, so a sentence with
clearly more of its tokens in one list was probably written with it. For each sentence $s_i$ and
each list $G_j$, the detector measures this excess with a score $\lambda_{ij}$, a log-likelihood
ratio that rises with each token of $s_i$ in $G_j$ and falls with each token outside it. Formally,
\begin{equation}
  \lambda_{ij}=g_{ij}\log\frac{p_1}{\gamma}+(n_i-g_{ij})\log\frac{1-p_1}{1-\gamma},
\end{equation}
where $n_i$ is the number of tokens in $s_i$, $g_{ij}$ how many of them are in $G_j$, and $p_1$ the
share we expect in $G_j$ if $s_i$ was written with it.

The detector does not need to know which list was active. It combines the scores of all lists,
giving more weight to the one the grammar of the previous sentence points to:
\begin{equation}
  S_i=\log\sum_{j=1}^{k} w_j\,e^{\lambda_{ij}},
\end{equation}
where the weight $w_j$ decreases with the distance from the profile of $s_{i-1}$ to the centre of
cell $j$, at a rate set by a temperature $\tau$. If the previous sentence is intact, the weights
point to the right list; if an edit has changed its grammar, a watermarked sentence is still
detected when its tokens fit the right list clearly enough. Because every list is tested, the
detector never has to recompute the key exactly. The first sentence is not scored.

Finally, the sentence scores are summed into a score for the whole text. An attack rarely alters
every sentence: it changes some and leaves the others intact, and the changed ones add noise but
little evidence of the watermark. The detector therefore leaves out the fraction $\rho$ of
sentences with the lowest scores. With $I$ the set of sentences that remain, the score of the text
is
\begin{equation}
  S=\sum_{i\in I} S_i .
\end{equation}
The threshold is set on unwatermarked text scored in the same way, so leaving sentences out does
not raise the false-positive rate.

\section{Attacks Adapted to the Published Design}\label{sec:threat}\label{sec:protocol}

\begin{widetable}[t]
  \centering
  \caption{Judged by its weaker axis, every watermark fails, and \KL outperforms them all}
  \label{tab:worst}
  \tablewithnote{\begin{tabular}{lcccccccc}
  \toprule
  & \multicolumn{3}{c}{OPT-1.3B} & \multicolumn{3}{c}{Mistral-7B} & \multicolumn{2}{c}{No attack (OPT)} \\
  \cmidrule(lr){2-4}\cmidrule(lr){5-7}\cmidrule(lr){8-9}
  Watermark & Forged & Erased & Weaker axis & Forged & Erased & Weaker axis & Detected & $\Delta$PPL \\
  \midrule
  SynthID \citep{dathathri2024synthid} & 0.06 & 0.66 & 0.66\textsuperscript{E} & 0.09 & 0.68 & 0.68\textsuperscript{E} & 1.000 & 15.12 \\
  DiPmark \citep{wu2024dipmark}        & 0.09 & 0.91 & 0.91\textsuperscript{E} & 0.08 & 0.92 & 0.92\textsuperscript{E} & 0.925 & 3.42 \\
  KGW-4 \citep{kirchenbauer2023}       & 0.05 & 0.93 & 0.93\textsuperscript{E} & 0.07 & 0.80 & 0.80\textsuperscript{E} & 0.975 & 4.93 \\
  \addlinespace
  Unigram \citep{zhao2024}             & 1.00 & 0.15 & 1.00\textsuperscript{F} & 1.00 & 0.18 & 1.00\textsuperscript{F} & 0.980 & 4.59 \\
  KGW-1 \citep{kirchenbauer2023}       & 0.94 & 0.37 & 0.94\textsuperscript{F} & 0.90 & 0.41 & 0.90\textsuperscript{F} & 0.990 & 4.80 \\
  SIR \citep{liu2024sir}               & 0.87 & 0.25 & 0.87\textsuperscript{F} & 0.94 & 0.22 & 0.94\textsuperscript{F} & 0.867 & 5.30 \\
  \addlinespace
  SemStamp \citep{hou2024semstamp}     & 0.96 & 1.00 & 1.00\textsuperscript{E} & 0.97 & n/a  & $\geq$0.97\textsuperscript{F} & 0.835 & $-$0.83 \\
  k-SemStamp \citep{hou2024ksemstamp}  & 0.97 & 1.00 & 1.00\textsuperscript{E} & 0.99 & n/a  & $\geq$0.99\textsuperscript{F} & 0.680 & $-$0.01 \\
  \midrule
  \oursrow \KL (ours)                  & 0.33 & 0.53 & \textbf{0.53}\textsuperscript{E} & 0.56 & 0.60 & \textbf{0.60}\textsuperscript{E} & 0.875 & 4.09 \\
  \bottomrule
\end{tabular}
}{Forged: how often the detector accepts the best forger's texts. Erased: how much of the signal the most damaging edit removes. Weaker axis: the worse of the two, marked E when erasure decides it and F when forgery does; lower is better. No attack: the watermarked text as generated. Detected: share of it flagged at a 1\% false-positive rate (higher is better). $\Delta$PPL: rise in perplexity, measured by an independent model, of watermarked over unwatermarked text from the same prompts; the cost in quality (lower is better). n/a: forgery already decides it. The results behind this table are in \cref{app:full}.}
\end{widetable}

\paragraph{The attacker.} We follow Kerckhoffs's principle \citep{kerckhoffs1883,jovanovic2024}:
the attacker knows how the watermark works, but not its key. This is not a pessimistic
assumption. It is what the providers have published, and where they have not, the design can be
recovered from the outside \citep{gloaguen2025blackbox}. The attacker has text watermarked by the
model, ordinary text from the same domain and a language model of its own. It has no access to the
detector, so it cannot see how its own text would be judged.

\paragraph{How it adapts.} Every watermark chooses its green list from something in the text, which
we call its \emph{context}: the previous word, the previous few words, the meaning of the sentence
or, for \KL, the grammar of the previous sentence. An attacker who knows the design knows what the
context is, and attacks through it on both axes (\cref{tab:attack-families}):
\begin{itemize}
  \item \textbf{To forge}, it reads the watermarked texts and, for each context, notes which words
  appear more often than in ordinary text: those are the words the key favours there. It then
  writes text of its own that leans towards them, so that the detector takes it as watermarked.
  \item \textbf{To erase}, it edits the watermarked text so that the context changes. Whether the
  context is the preceding words or the sentence, an attacker can employ rewording in the former
  and modify punctuation in the latter.
\end{itemize}

\paragraph{How we measure.} To measure forgery, we let each attacker write text and count how
often the genuine detector accepts it, at the same 5\% false-positive rate for every watermark. As
a control, the same attacker also pushes towards a random list instead of the one it learned; what
it gains over that control is what it actually learned about the key. To measure erasure, we use the
AUROC of the detector after the attack, as earlier work does \citep{kirchenbauer2024reliability}.
We apply the edit to watermarked and ordinary text alike and report how much of the detector's
ability to tell them apart survives, from all of it to none. Every watermark is attacked on two generators,
OPT-1.3B and Mistral-7B. \Cref{app:measure} gives the details.

\paragraph{The weaker axis.} A watermark is only as strong as its weaker axis. For each watermark
we take the best forgery and the largest share of the signal removed by any attack, adapted or
not, such as general paraphrasers and distillation, and score it by whichever is worse. A
watermark that holds on one axis but gives way on the other has not held. The two axes are
measured on different scales, a share of forgeries accepted and a share of signal lost, so we
always state which of the two is the weaker axis.

\section{Results}\label{sec:results}

\paragraph{Every watermark fails on one axis, and \KL outperforms them all.} Measured one axis at a
time, every existing watermark looks secure on one of them, and this gives a false sense of
security: an attacker who reads the published design attacks the other axis, and there every
existing watermark fails (\cref{tab:worst}; \cref{app:full} gives the full breakdown). \KL, on the
other hand, has no such weak axis. It is not the best on either axis alone, but its weaker axis is
stronger than that of any other watermark, on both generators. It loses less signal than the
watermarks that, like it, are weaker against erasure, and it is forged far less often than those
weaker against forgery. What defeats the others is not a
stronger attack but an adapted one.

\subsection{Why the existing watermarks fail}\label{sec:forgery}

Each existing watermark fails on the axis its context exposes. A key read from the previous few
tokens, as in SynthID, DiPmark and KGW-4, almost never repeats, so the attacker cannot learn it and
forge it; but changing one word changes the key, so paraphrase erases it, as the providers
themselves report \citep{anthropic2026watermark,openai2026provenance}. A key read from something
that rewording barely disturbs, a fixed list in Unigram, the previous word in KGW-1 or the meaning
of the text in SIR, survives paraphrase; but the attacker sees each list many times and forges it.
SemStamp and k-SemStamp fail on both: their few regions of meaning are easy to learn, and removing
the punctuation erases their mark.

\subsection{Why \KL outperforms the rest}\label{sec:robustness}

\KL handles this trade-off better by separating what the attacker can see from what it needs. The
grammar of each sentence is public, so the attacker knows when the list changes, but only the key
decides what is in each list, and the attacker has to learn every list word by word. \KL is
therefore forged less often than every watermark that survives paraphrase, in every setting and
against every attacker we ran.

On erasure, the detector tries every list instead of recomputing the key, so edits that leave most
of each sentence in place barely touch the mark. Paraphrase costs it more, but it still loses less
of its signal than every watermark that resists forgery. What hurts it most is the adaptive
paraphraser and removing the punctuation (\cref{tab:price}).

These results hold with another generator, another domain, a gentler push and stronger attackers
(\cref{app:full}). Without attack, \KL is detected about as well as SIR, at no extra cost in
text quality over the best-known watermarks, and tagging each sentence barely slows generation
(\cref{app:k}). \Cref{sec:mechanism} shows which design choice gives each of these properties.

\section{Taking \KL Apart: One Design Choice per Attack}\label{sec:mechanism}

\KL makes several design choices at once: its key is read from grammar, it changes once per
sentence, and its detector tries every list instead of recomputing the key. To see which choice
gives which property, we change them one at a time (\cref{tab:mechanism}).

\begin{table}[!htb]
  \centering
  \caption{Against counting, the text must be read sentence by sentence; against edits, the
  detector must try every list}
  \label{tab:mechanism}
  \tablewithnote{
\setlength{\tabcolsep}{4pt}%
\begin{tabular}{lccc}
  \toprule
  & & \multicolumn{2}{c}{Signal kept} \\
  \cmidrule(lr){3-4}
  Change to \KL & Forged & DIPPER & Synonyms \\
  \midrule
  \oursrow None (\KL)                  & 0.30 & 0.62 & 0.93 \\
  Key from a word hash          & 0.34 & 0.39 & 0.80 \\
  Key every 3 tokens            & 0.37 & 0.49 & 0.97 \\
  20-token windows              & 0.54 & 0.56 & 0.90 \\
  Key ignored                   & 0.35 & 0.57 & 0.90 \\
  Word hash, key ignored        & 0.40 & 0.58 & 0.90 \\
  \bottomrule
\end{tabular}
}{Each row is \KL with one design choice changed; the last row changes two. Forged: how often the detector accepts the forger's texts. Signal kept under DIPPER and synonym substitution. OPT-1.3B.}
\end{table}

\paragraph{Against counting, the text must be read sentence by sentence.} The more often the key
changes, the easier it is to forge: KGW-1, whose key changes with every word, is forged most of
the time (\cref{tab:worst}), while \KL, whose key changes once per sentence, is forged about a
third of the time. Where the key comes from makes no difference: a key taken from the words of the
sentence does as well as grammar. The detector, however, does: if it reads the text in blocks of
20 tokens instead of sentence by sentence, forgeries get through about half of the time.

\paragraph{Against edits, the detector must try every list.} An edit can change the key: if the
key is read from the words of the previous sentence, rewording that sentence points to the wrong
list, and a detector that follows the key looks in the wrong place. But the watermarked words are
still there. Our detector does not depend on the key: it checks each sentence against every list,
and the right one stands out. Used this way, a key taken from the words survives edits as well as
grammar does. This is what lets \KL be hard to forge and robust at the same time. KGW-1 cannot do
the same: its list changes with every word, and a single word is too little to tell which list it
came from.

\paragraph{Publishing the design, and averaging over keys.} Telling the forger how \KL groups
sentences barely helps it (\cref{tab:attackers-kl}), so the whole design can be published; only the lists must stay secret.
Grammar matters against one more attack, averaging over keys \citep{pang2024}, with full results
in \cref{app:grammar}: the attacker rewords the previous sentence again and
again to reach other lists, and averages over them until the mark fades. Rewording changes the
words of a sentence almost every time, but its grammar much less often, so most texts are still
detected. In short, each attack is stopped by a different choice: reading sentence by sentence
stops counting, the detector stops edits, and grammar stops averaging over keys. None of them
needs the design to be secret.

\section{Conclusion}\label{sec:conclusion}\label{sec:limitations}

A watermark that is easy to steal lets anyone put words in a model's mouth, and a watermark that
breaks under paraphrase lets anyone take them out. Providers now deploy watermarks and explain how
they work, so an attacker knows the design, and the design tells it which of the two attacks to
use. Taken one axis at a time, every kind of watermark seems to have a safe choice; judged by its
weaker axis, every existing watermark we compared fails. The family the providers deploy cannot be
forged but is erased by paraphrase, and the watermarks that survive paraphrase are forged most of
the time.

\KL does not escape this trade-off, but it outperforms every watermark we compare, on both generators,
at no extra cost in text quality over the best-known watermarks. Taking it apart gives a simple recipe for a
watermark that can be published. Keep only the lists secret, and publish everything else. Read the
text sentence by sentence, so that an attacker cannot learn the lists by counting. Let the
detector try every list, so that an edit that changes the key does not erase the mark. And read
the key from grammar, which paraphrasing alters less than it alters words.

Our comparison has a natural scope: it uses SynthID as
released, since the versions Anthropic and OpenAI deploy are not public, and English text from two
generators.

A key takeaway for future work on watermark attacks is to attack each watermark through what its
design discloses, and to report its weaker axis, and which axis it is, rather than each axis apart. For deployment, the
family in use today resists forgery but is erased by a paraphrase, as the providers themselves
report. Separating sentences without relying on punctuation, and carrying \KL to other languages,
are the next steps.

\onecolumn
\appendices
\crefalias{section}{appendix}
\raggedbottom

\begin{multicols}{2}
\section{Setup and Measurement}\label{app:measure}\label{app:repro}

\paragraph{Generation.} The generators are OPT-1.3B \citep{zhang2022opt}, in half precision, and
Mistral-7B \citep{jiang2023mistral}, in bfloat16. Prompts are inputs of C4-RealNewsLike documents
\citep{raffel2020c4} and, as a second domain, of BookSum \citep{kryscinski2022booksum}. A
watermarked text and its unwatermarked twin share the same random seed, and the detector scores
only what the model wrote. Baselines run with their published settings through MarkLLM
\citep{pan2024markllm}, with the same decoding as \KL. \Cref{tab:params} lists the parameters of
\KL; only $k$ and $\rho$ were chosen by measurement.

\begin{apptable}
  \caption{Parameters of \KL}
  \label{tab:params}
  \setlength{\tabcolsep}{4pt}%
\begin{tabularx}{\linewidth}{@{}ll>{\raggedright\arraybackslash}X@{}}
  \toprule
  Parameter & Value & What it does; how it was set \\
  \midrule
  Cells $k$ & 4 & Groups, one list each; swept from 2 to 16 \\
  Trimmed share $\rho$ & 0.4 & Worst sentences set aside; compared with no trimming \\
  Bias $\delta$ & 2 & Push towards the green list; usual value, not tuned \\
  Green share $\gamma$ & 0.5 & Vocabulary in each list; usual value, not tuned \\
  Temperature $\tau$ & 0.3 & Trust in the grammatical hint; starting value, not tuned \\
  Tagger & spaCy 3.8 & Labels word classes; small English model (\texttt{en\_core\_web\_sm}) \\
  Fitting text & 20 C4 docs & Fits grouping and detector; disjoint from every prompt \\
  \bottomrule
\end{tabularx}

\end{apptable}

\paragraph{Forgery.} A forger wins when the genuine detector, set to a 5\% false-positive rate,
takes its text for watermarked; we report how often it wins over 600 texts of 200 tokens. As a
control, the same forger also pushes towards a random list, and a result whose control wins more
than 35\% of the time is not used. There are three forgers: the counting attacker of
\cref{sec:threat}, which learns a list per context from 200 watermarked texts; the attacker of
\citet{jovanovic2024}, which combines several contexts over 5\,000 to 12\,000 texts; and
distillation \citep{gu2024learnability}, which fine-tunes a copy of the model on 10\,000
watermarked texts.

\paragraph{Erasure.} The detector must tell watermarked text from ordinary text; its AUROC is 1
when it always does and 0.5 when it guesses. We apply each edit to 200 watermarked and 200
unwatermarked texts and report \emph{retention}, the part of the detector's advantage over
guessing that survives:
\[ \text{retention}=\frac{\mathrm{AUROC}_{\text{after}}-0.5}{\mathrm{AUROC}_{\text{before}}-0.5}, \]
1 when the edit changes nothing and 0 when the mark is gone. The signal erased in the body is one
minus retention. Three edits decide the results: DIPPER \citep{krishna2023}, a paraphraser trained
to evade detectors; the adaptive paraphraser of \citet{diaa2025adaptive}, tuned against Unigram
and used by transfer against the others; and removing all punctuation. We also ran
back-translation, a T5 paraphraser, and five light edits (synonyms, case, typos, shuffled words
and an inserted passage).

\paragraph{Code and hardware.} The implementation, a script per table and the result files of
every run will be released upon acceptance. Runs used NVIDIA L40S and A100 GPUs; re-runs on other
hardware are close but not identical.

\section{Full Results}\label{app:full}

\Cref{tab:worst} keeps one number per watermark and axis: the worst. \Cref{tab:generalise,tab:price}
give the numbers it was chosen from, with the one it keeps in bold.

\paragraph{Forgery.} Each column of \cref{tab:generalise} is a forger, and each number how often
its texts pass as watermarked; in parentheses, its control. The pattern is the same in every
column: the watermarks that read their key from the previous few tokens are almost never forged, those with a stable key are forged most of the time, and \KL is in between. Its
highest rate on OPT-1.3B, 0.33, comes from a forger given the defender's grouping
(\cref{tab:attackers-kl}). On BookSum and with a gentler push, the counting attacker gives the
same picture: \KL is forged 0.28 and 0.33 of the time, the watermarks with a stable key from 0.44
to 1.00, and the others at most 0.06.

\end{multicols}
\begin{apptable}
  \caption{Forgery rate against every forger}
  \label{tab:generalise}\label{tab:security}
  \tablewithnote{\begin{tabular}{lcccccc}
  \toprule
  & \multicolumn{3}{c}{OPT-1.3B} & \multicolumn{3}{c}{Mistral-7B} \\
  \cmidrule(lr){2-4}\cmidrule(lr){5-7}
  Watermark & Counting & Distillation & Jovanovi\'c & Counting & Distillation & Jovanovi\'c \\
  \midrule
  SynthID & 0.05 (0.05) & 0.05 & \textbf{0.06} & 0.06 (0.05) & \textbf{0.09} & \textbf{0.09} \\
  DiPmark & 0.05 (0.05) & 0.06 & \textbf{0.09} & 0.05 (0.05) & 0.05 & \textbf{0.08} \\
  KGW-4 & \textbf{0.05} (0.05) & -- & -- & \textbf{0.07} (0.05) & -- & -- \\
  \addlinespace
  Unigram & 0.97 (0.43) & 0.97 & \textbf{1.00} & 0.89 (0.05) & \textbf{1.00} & 0.98 \\
  KGW-1 & 0.81 (0.07) & 0.57 & \textbf{0.94} & 0.55 (0.09) & 0.83 & \textbf{0.90} \\
  SIR & 0.76 (0.15) & \textbf{0.87} & 0.81 & 0.51 (0.07) & \textbf{0.94} & 0.59 \\
  \addlinespace
  SemStamp & \textbf{0.96} (0.01) & -- & -- & \textbf{0.97} (0.02) & -- & -- \\
  k-SemStamp & \textbf{0.97} (0.00) & -- & -- & \textbf{0.99} (0.01) & -- & -- \\
  \midrule
  \oursrow \KL & 0.30 (0.14) & 0.22 & 0.22 & 0.24 (0.09) & \textbf{0.56} & 0.08 \\
  \bottomrule
\end{tabular}
}{In parentheses, the control: the same forger pushing towards a random list. In bold, the rate that enters \cref{tab:worst}. --: not run. SemStamp and k-SemStamp are already forged at least 96\% of the time by the counting attacker, so a stronger forger cannot change their result. KGW-4 reads its key the same way as SynthID and DiPmark, which stay at the level of the control against the stronger forgers, so they were not run against it.}
\end{apptable}
\begin{apptable}
  \caption{Signal kept under the edits that decide the results}
  \label{tab:price}
  \tablewithnote{\begin{tabular}{lcccccccc}
  \toprule
  & \multicolumn{4}{c}{OPT-1.3B} & \multicolumn{4}{c}{Mistral-7B} \\
  \cmidrule(lr){2-5}\cmidrule(lr){6-9}
  Watermark & DIPPER & Adaptive & Punct. & Other & DIPPER & Adaptive & Punct. & Other \\
  \midrule
  SynthID & 0.43 & \textbf{0.34} & 0.97 & 0.55 & \textbf{0.32} & 0.37 & 0.99 & 0.62 \\
  DiPmark & 0.21 & \textbf{0.09} & 0.90 & 0.31 & \textbf{0.08} & 0.22 & 0.92 & 0.34 \\
  KGW-4 & 0.21 & \textbf{0.07} & 0.93 & 0.36 & 0.22 & \textbf{0.20} & 0.97 & 0.46 \\
  \addlinespace
  Unigram & 0.92 & \textbf{0.85} & 0.99 & 0.96 & 0.93 & \textbf{0.82} & 1.00 & 0.92 \\
  KGW-1 & 0.82 & \textbf{0.63} & 0.99 & 0.89 & 0.72 & \textbf{0.59} & 1.00 & 0.89 \\
  SIR & \textbf{0.76} & 0.77 & 0.99 & 0.87 & 0.81 & \textbf{0.78} & 0.99 & 0.88 \\
  \addlinespace
  SemStamp & 0.54 & -- & \textbf{0.00} & 0.62 & -- & -- & -- & -- \\
  k-SemStamp & 0.54 & -- & \textbf{0.00} & 0.63 & -- & -- & -- & -- \\
  \midrule
  \oursrow \KL & 0.62 & \textbf{0.47} & 0.52 & 0.76 & 0.49 & \textbf{0.40} & 0.47 & 0.72 \\
  \quad 20-token windows & 0.56 & -- & 0.96 & 0.76 & -- & -- & -- & -- \\
  \bottomrule
\end{tabular}
}{Retention, from 1 (all kept) to 0 (all lost), C4. Other: the worst of the seven remaining edits. In bold, the value behind \cref{tab:worst}. --: not run. The adaptive paraphraser was run only where it could change \cref{tab:worst}: not on SemStamp and k-SemStamp, which removing the punctuation already erases, and which are not attacked on Mistral-7B, where forgery already decides their weaker axis. The 20-token variant of \KL is not in \cref{tab:worst} and was measured on OPT-1.3B only.}
\end{apptable}
\begin{multicols}{2}

\paragraph{Erasure.} One minus the bold value in \cref{tab:price} is the signal erased in
\cref{tab:worst}. The paraphrasers separate the watermarks: the ones that read their key from the
previous few tokens lose most of their signal, those with a stable key keep most of it, and \KL is
in between.
Removing the punctuation erases the sentence-level watermarks and costs \KL about half of its
signal; reading the text in 20-token windows instead of sentences closes that gap.

\section{Attacking \KL with Knowledge of Its Design}\label{app:kl-attacks}\label{app:grammar}

\begin{apptable}
  \caption{Knowing the design of \KL barely helps the forger}
  \label{tab:attackers-kl}
  \tablewithnote{\begin{tabular}{lc}
  \toprule
  The forger & Forged \\
  \midrule
  \oursrow groups sentences itself, from 20 documents  & 0.30 \\
  groups sentences itself, from 200 documents         & 0.24 \\
  discovers the grouping by EM                        & 0.25 \\
  is given the defender's grouping                    & \textbf{0.33} \\
  \quad and reads 1\,000 texts instead of 200         & 0.22 \\
  writes every sentence with one list                 & 0.21 \\
  \quad with the defender's grouping                  & 0.22 \\
  \addlinespace
  is given the true lists                             & 0.56 \\
  \bottomrule
\end{tabular}
}{Forgery rate, OPT-1.3B, C4. In bold, the rate that enters \cref{tab:worst}.}
\end{apptable}

\paragraph{The grouping.} No attack on the grouping takes \KL above a third
(\cref{tab:attackers-kl}). Discovering the grouping, being given it, reading five times more text
or writing every sentence with a single list all stay between 0.21 and 0.33. Even an
attacker handed the true lists passes only about half of the time.

\paragraph{The grammar.} At the strength we use, reading the list from the grammar of the previous
sentence helps the detector only a little (\cref{tab:mechanism}). It matters in two cases. With a
weaker watermark, knowing which list to expect lets the detector find a mark that no longer stands
out, raising detection from 0.26 to 0.43; reading the list from the words does as well on
untouched text, but after synonym substitution it keeps 0.53 of the signal, against 0.84 with
grammar. Against averaging over keys, the attacker rewords the previous sentence to reach other
lists and mixes them; since rewording changes the grammar much less often than the words, it
reaches about two lists instead of four, and 0.59 of the texts are still detected, against 0.27
when the list is read from the words.

\section{Choice of \texorpdfstring{$k$}{k} and Cost}\label{app:k}

\paragraph{Number of cells.} With two cells the lists are so few that even a forger pushing at
random passes 56\% of the time. With four, eight and sixteen, forgery stays near 25\%, while the
detector finds 92\% of its own text with four or eight cells and 85\% with sixteen; we use four
(OPT-1.3B, C4).

\paragraph{Cost.} On OPT-1.3B and one GPU, tagging each finished sentence slows generation by 7\%,
against 24\% for SynthID and a factor of three for SIR, which embeds the preceding text at every
step. Detecting a 200-token text takes 18\,ms, against 9 to 69\,ms for the other watermarks and
262\,ms for SIR.
\end{multicols}

\twocolumn
\bibliographystyle{IEEEtranN}
\bibliography{references}

\begin{thebibliography}{40}
\providecommand{\natexlab}[1]{#1}
\providecommand{\url}[1]{#1}
\csname url@samestyle\endcsname
\providecommand{\newblock}{\relax}
\providecommand{\bibinfo}[2]{#2}
\providecommand{\BIBentrySTDinterwordspacing}{\spaceskip=0pt\relax}
\providecommand{\BIBentryALTinterwordstretchfactor}{4}
\providecommand{\BIBentryALTinterwordspacing}{\spaceskip=\fontdimen2\font plus
\BIBentryALTinterwordstretchfactor\fontdimen3\font minus \fontdimen4\font\relax}
\providecommand{\BIBforeignlanguage}[2]{{%
\expandafter\ifx\csname l@#1\endcsname\relax
\typeout{** WARNING: IEEEtranN.bst: No hyphenation pattern has been}%
\typeout{** loaded for the language `#1'. Using the pattern for}%
\typeout{** the default language instead.}%
\else
\language=\csname l@#1\endcsname
\fi
#2}}
\providecommand{\BIBdecl}{\relax}
\BIBdecl

\bibitem[{European Parliament and Council of the European Union}(2024)]{euaiact2024}
{European Parliament and Council of the European Union}, ``Regulation ({EU}) 2024/1689 laying down harmonised rules on artificial intelligence ({Artificial Intelligence Act}),'' \url{https://eur-lex.europa.eu/eli/reg/2024/1689/oj}, 2024, official Journal of the European Union, L, 12 July 2024; Article 50(2).

\bibitem[Shumailov et~al.(2024)Shumailov, Shumaylov, Zhao, Papernot, Anderson, and Gal]{shumailov2024collapse}
I.~Shumailov, Z.~Shumaylov, Y.~Zhao, N.~Papernot, R.~Anderson, and Y.~Gal, ``{AI} models collapse when trained on recursively generated data,'' \emph{Nature}, vol. 631, pp. 755--759, 2024.

\bibitem[Sadasivan et~al.(2025)Sadasivan, Kumar, Balasubramanian, Wang, and Feizi]{sadasivan2023}
V.~S. Sadasivan, A.~Kumar, S.~Balasubramanian, W.~Wang, and S.~Feizi, ``Can {AI}-generated text be reliably detected? stress testing {AI} text detectors under various attacks,'' \emph{Transactions on Machine Learning Research}, 2025.

\bibitem[Dathathri et~al.(2024)Dathathri, See, Ghaisas, Huang, McAdam, et~al.]{dathathri2024synthid}
S.~Dathathri, A.~See, S.~Ghaisas, P.-S. Huang, R.~McAdam \emph{et~al.}, ``Scalable watermarking for identifying large language model outputs,'' \emph{Nature}, vol. 634, no. 8035, pp. 818--823, 2024.

\bibitem[{Anthropic}(2026)]{anthropic2026watermark}
{Anthropic}, ``How {Claude}'s text watermark works,'' \url{https://www.anthropic.com/news/claude-text-watermark}, Aug. 2026, published 14 August 2026, updated 1 September 2026.

\bibitem[Li et~al.(2026)Li, Wen, Chen, Long, Jain, Joly, Lam, Song, and Su]{li2026textgrain}
X.~Li, G.~Wen, X.~Chen, Q.~Long, A.~Jain, F.~Joly, M.~Lam, Q.~Song, and W.~Su, ``{textGrain}: Entropy-calibrated watermarking for language model text,'' OpenAI, Tech. Rep., Oct. 2026, technical report accompanying OpenAI's 5 October 2026 blog post.

\bibitem[Kirchenbauer et~al.(2023)Kirchenbauer, Geiping, Wen, Katz, Miers, and Goldstein]{kirchenbauer2023}
J.~Kirchenbauer, J.~Geiping, Y.~Wen, J.~Katz, I.~Miers, and T.~Goldstein, ``A watermark for large language models,'' in \emph{Proceedings of the 40th International Conference on Machine Learning}, ser. ICML, vol. 202, 2023, pp. 17\,061--17\,084.

\bibitem[Liu et~al.(2024{\natexlab{a}})Liu, Pan, Lu, Li, Hu, Zhang, Wen, King, Xiong, and Yu]{liu2024survey}
A.~Liu, L.~Pan, Y.~Lu, J.~Li, X.~Hu, X.~Zhang, L.~Wen, I.~King, H.~Xiong, and P.~S. Yu, ``A survey of text watermarking in the era of large language models,'' arXiv:2312.07913, 2024.

\bibitem[Zhao et~al.(2024)Zhao, Ananth, Li, and Wang]{zhao2024}
X.~Zhao, P.~Ananth, L.~Li, and Y.-X. Wang, ``Provable robust watermarking for {AI}-generated text,'' in \emph{Proceedings of the 12th International Conference on Learning Representations}, ser. ICLR, 2024.

\bibitem[Liu et~al.(2024{\natexlab{b}})Liu, Pan, Hu, Meng, and Wen]{liu2024sir}
A.~Liu, L.~Pan, X.~Hu, S.~Meng, and L.~Wen, ``A semantic invariant robust watermark for large language models,'' in \emph{Proceedings of the 12th International Conference on Learning Representations}, ser. ICLR, 2024.

\bibitem[Wu et~al.(2024)Wu, Hu, Guo, Zhang, and Huang]{wu2024dipmark}
\BIBentryALTinterwordspacing
Y.~Wu, Z.~Hu, J.~Guo, H.~Zhang, and H.~Huang, ``A resilient and accessible distribution-preserving watermark for large language models,'' in \emph{Proceedings of the 41st International Conference on Machine Learning}, ser. Proceedings of Machine Learning Research, vol. 235, 2024, pp. 53\,443--53\,470. [Online]. Available: \url{https://proceedings.mlr.press/v235/wu24h.html}
\BIBentrySTDinterwordspacing

\bibitem[Kuditipudi et~al.(2024)Kuditipudi, Thickstun, Hashimoto, and Liang]{kuditipudi2024robust}
R.~Kuditipudi, J.~Thickstun, T.~Hashimoto, and P.~Liang, ``Robust distortion-free watermarks for language models,'' \emph{Transactions on Machine Learning Research}, 2024.

\bibitem[Hou et~al.(2024{\natexlab{a}})Hou, Zhang, He, Wang, Chuang, Wang, Shen, Van~Durme, Khashabi, and Tsvetkov]{hou2024semstamp}
A.~B. Hou, J.~Zhang, T.~He, Y.~Wang, Y.-S. Chuang, H.~Wang, L.~Shen, B.~Van~Durme, D.~Khashabi, and Y.~Tsvetkov, ``{SemStamp}: A semantic watermark with paraphrastic robustness for text generation,'' in \emph{Proceedings of the 2024 Conference of the North American Chapter of the Association for Computational Linguistics: Human Language Technologies (Volume 1: Long Papers)}, 2024, pp. 4067--4082.

\bibitem[Jovanovi{\'c} et~al.(2024)Jovanovi{\'c}, Staab, and Vechev]{jovanovic2024}
N.~Jovanovi{\'c}, R.~Staab, and M.~Vechev, ``Watermark stealing in large language models,'' in \emph{Proceedings of the 41st International Conference on Machine Learning}, ser. ICML, vol. 235, 2024, pp. 22\,570--22\,593.

\bibitem[Krishna et~al.(2023)Krishna, Song, Karpinska, Wieting, and Iyyer]{krishna2023}
K.~Krishna, Y.~Song, M.~Karpinska, J.~Wieting, and M.~Iyyer, ``Paraphrasing evades detectors of {AI}-generated text, but retrieval is an effective defense,'' in \emph{Advances in Neural Information Processing Systems}, ser. NeurIPS, 2023.

\bibitem[Diaa et~al.(2025)Diaa, Aremu, and Lukas]{diaa2025adaptive}
A.~Diaa, T.~Aremu, and N.~Lukas, ``Optimizing adaptive attacks against watermarks for language models,'' in \emph{Proceedings of the 42nd International Conference on Machine Learning (ICML)}, ser. PMLR, vol. 267, 2025, pp. 13\,546--13\,569.

\bibitem[Huang et~al.(2025)Huang, Zhang, Zheng, Gong, Li, Liu, Liu, and Liang]{huang2025rlcracker}
H.~Huang, Y.~Zhang, H.~Zheng, X.~Gong, Y.~Li, L.~Liu, Z.~Liu, and S.~Liang, ``{RLCracker}: Evaluating the worst-case vulnerability of {LLM} watermarks with adaptive {RL} attacks,'' \emph{arXiv preprint arXiv:2509.20924}, 2025.

\bibitem[Zhang et~al.(2026)Zhang, Zhang, Zhang, Zhang, Pan, Liu, Gill, and Zhang]{zhang2026character}
Z.~Zhang, X.~Zhang, Y.~Zhang, H.~Zhang, S.~Pan, B.~Liu, A.~Q. Gill, and L.~Y. Zhang, ``Character-level perturbations disrupt {LLM} watermarks,'' in \emph{Network and Distributed System Security Symposium (NDSS)}, 2026.

\bibitem[Gu et~al.(2024)Gu, Li, Liang, and Hashimoto]{gu2024learnability}
C.~Gu, X.~L. Li, P.~Liang, and T.~Hashimoto, ``On the learnability of watermarks for language models,'' in \emph{The Twelfth International Conference on Learning Representations (ICLR)}, 2024.

\bibitem[Gloaguen et~al.(2025)Gloaguen, Jovanovi{\'c}, Staab, and Vechev]{gloaguen2025blackbox}
T.~Gloaguen, N.~Jovanovi{\'c}, R.~Staab, and M.~Vechev, ``Black-box detection of language model watermarks,'' in \emph{International Conference on Learning Representations}, ser. ICLR, 2025.

\bibitem[Kerckhoffs(1883)]{kerckhoffs1883}
A.~Kerckhoffs, ``La cryptographie militaire,'' \emph{Journal des sciences militaires}, vol.~9, pp. 5--38, 1883.

\bibitem[Wang et~al.(2025)Wang, Gu, Wu, and Yang]{wang2025morphmark}
Z.~Wang, T.~Gu, B.~Wu, and Y.~Yang, ``{MorphMark}: Flexible adaptive watermarking for large language models,'' in \emph{Proceedings of the 63rd Annual Meeting of the Association for Computational Linguistics (Volume 1: Long Papers)}.\hskip 1em plus 0.5em minus 0.4em\relax Vienna, Austria: Association for Computational Linguistics, 2025, pp. 4842--4860.

\bibitem[Fernandez et~al.(2023)Fernandez, Chaffin, Tit, Chappelier, and Furon]{fernandez2023threebricks}
P.~Fernandez, A.~Chaffin, K.~Tit, V.~Chappelier, and T.~Furon, ``Three bricks to consolidate watermarks for large language models,'' in \emph{IEEE International Workshop on Information Forensics and Security (WIFS)}, 2023.

\bibitem[He et~al.(2024)He, Zhou, Hao, Liu, Wang, Tu, Zhang, and Wang]{he2024xsir}
Z.~He, B.~Zhou, H.~Hao, A.~Liu, X.~Wang, Z.~Tu, Z.~Zhang, and R.~Wang, ``Can watermarks survive translation? on the cross-lingual consistency of text watermark for large language models,'' in \emph{Proceedings of the 62nd Annual Meeting of the Association for Computational Linguistics (Volume 1: Long Papers)}, 2024, pp. 4115--4129.

\bibitem[Liu and Bu(2024)]{liu2024adaptive}
Y.~Liu and Y.~Bu, ``Adaptive text watermark for large language models,'' in \emph{Proceedings of the 41st International Conference on Machine Learning}, ser. Proceedings of Machine Learning Research, vol. 235, 2024, pp. 30\,718--30\,737.

\bibitem[Chang et~al.(2024)Chang, Krishna, Houmansadr, Wieting, and Iyyer]{chang2024postmark}
Y.~Chang, K.~Krishna, A.~Houmansadr, J.~Wieting, and M.~Iyyer, ``{PostMark}: A robust blackbox watermark for large language models,'' in \emph{Proceedings of the 2024 Conference on Empirical Methods in Natural Language Processing}.\hskip 1em plus 0.5em minus 0.4em\relax Miami, Florida, USA: Association for Computational Linguistics, 2024, pp. 8969--8987.

\bibitem[Hou et~al.(2024{\natexlab{b}})Hou, Zhang, Wang, Khashabi, and He]{hou2024ksemstamp}
A.~B. Hou, J.~Zhang, Y.~Wang, D.~Khashabi, and T.~He, ``k-{SemStamp}: A clustering-based semantic watermark for detection of machine-generated text,'' in \emph{Findings of the Association for Computational Linguistics: ACL 2024}, 2024, pp. 1706--1715.

\bibitem[Piet et~al.(2023)Piet, Sitawarin, Fang, Mu, and Wagner]{piet2023markmywords}
J.~Piet, C.~Sitawarin, V.~Fang, N.~Mu, and D.~Wagner, ``Mark my words: Analyzing and evaluating language model watermarks,'' 2023.

\bibitem[Pang et~al.(2024)Pang, Hu, Zheng, and Smith]{pang2024}
Q.~Pang, S.~Hu, W.~Zheng, and V.~Smith, ``No free lunch in {LLM} watermarking: Trade-offs in watermarking design choices,'' in \emph{Advances in Neural Information Processing Systems}, ser. NeurIPS, 2024.

\bibitem[Kirchenbauer et~al.(2024)Kirchenbauer, Geiping, Wen, Shu, Saifullah, Kong, Fernando, Saha, Goldblum, and Goldstein]{kirchenbauer2024reliability}
\BIBentryALTinterwordspacing
J.~Kirchenbauer, J.~Geiping, Y.~Wen, M.~Shu, K.~Saifullah, K.~Kong, K.~Fernando, A.~Saha, M.~Goldblum, and T.~Goldstein, ``On the reliability of watermarks for large language models,'' in \emph{The Twelfth International Conference on Learning Representations (ICLR)}, 2024. [Online]. Available: \url{https://openreview.net/forum?id=DEJIDCmWOz}
\BIBentrySTDinterwordspacing

\bibitem[Shen et~al.(2025)Shen, Huang, and Wan]{shen2025seek}
H.~Shen, B.~Huang, and X.~Wan, ``Enhancing {LLM} watermark resilience against both scrubbing and spoofing attacks,'' in \emph{Advances in Neural Information Processing Systems}, 2025.

\bibitem[An et~al.(2025)An, Liu, Liu, Zhang, Bu, and Chang]{an2025contrastive}
L.~An, Y.~Liu, Y.~Liu, Y.~Zhang, Y.~Bu, and S.~Chang, ``Defending {LLM} watermarking against spoofing attacks with contrastive representation learning,'' \emph{arXiv preprint arXiv:2504.06575}, 2025.

\bibitem[Liang et~al.(2025)Liang, Wang, Hong, Ji, and Wang]{liang2025waterpark}
J.~Liang, Z.~Wang, S.~Hong, S.~Ji, and T.~Wang, ``Watermark under fire: A robustness evaluation of {LLM} watermarking,'' in \emph{Findings of the Association for Computational Linguistics: EMNLP 2025}, 2025, pp. 21\,050--21\,074.

\bibitem[Nivre et~al.(2020)Nivre, de~Marneffe, Ginter, Haji{\v{c}}, Manning, Pyysalo, Schuster, Tyers, and Zeman]{nivre2020ud}
J.~Nivre, M.-C. de~Marneffe, F.~Ginter, J.~Haji{\v{c}}, C.~D. Manning, S.~Pyysalo, S.~Schuster, F.~Tyers, and D.~Zeman, ``Universal {D}ependencies v2: An evergrowing multilingual treebank collection,'' in \emph{Proceedings of the Twelfth Language Resources and Evaluation Conference}.\hskip 1em plus 0.5em minus 0.4em\relax Marseille, France: European Language Resources Association, 2020, pp. 4034--4043.

\bibitem[{OpenAI}(2026)]{openai2026provenance}
{OpenAI}, ``Our approach to {EU} text provenance rules,'' \url{https://openai.com/index/eu-text-provenance/}, Oct. 2026.

\bibitem[Zhang et~al.(2022)Zhang, Roller, Goyal, Artetxe, et~al.]{zhang2022opt}
S.~Zhang, S.~Roller, N.~Goyal, M.~Artetxe \emph{et~al.}, ``{OPT}: Open pre-trained transformer language models,'' \emph{arXiv preprint arXiv:2205.01068}, 2022.

\bibitem[Jiang et~al.(2023)Jiang, Sablayrolles, Mensch, et~al.]{jiang2023mistral}
A.~Q. Jiang, A.~Sablayrolles, A.~Mensch \emph{et~al.}, ``Mistral {7B},'' \emph{arXiv preprint arXiv:2310.06825}, 2023.

\bibitem[Raffel et~al.(2020)Raffel, Shazeer, Roberts, Lee, Narang, Matena, Zhou, Li, and Liu]{raffel2020c4}
C.~Raffel, N.~Shazeer, A.~Roberts, K.~Lee, S.~Narang, M.~Matena, Y.~Zhou, W.~Li, and P.~J. Liu, ``Exploring the limits of transfer learning with a unified text-to-text transformer,'' \emph{Journal of Machine Learning Research}, vol.~21, no. 140, pp. 1--67, 2020.

\bibitem[Kry{\'s}ci{\'n}ski et~al.(2022)Kry{\'s}ci{\'n}ski, Rajani, Agarwal, Xiong, and Radev]{kryscinski2022booksum}
W.~Kry{\'s}ci{\'n}ski, N.~Rajani, D.~Agarwal, C.~Xiong, and D.~Radev, ``{BookSum}: A collection of datasets for long-form narrative summarization,'' in \emph{Findings of the Association for Computational Linguistics: EMNLP 2022}, 2022, pp. 6536--6558.

\bibitem[Pan et~al.(2024)Pan, Liu, He, et~al.]{pan2024markllm}
L.~Pan, A.~Liu, Z.~He \emph{et~al.}, ``{MarkLLM}: An open-source toolkit for {LLM} watermarking,'' in \emph{Proceedings of EMNLP 2024: System Demonstrations}, 2024, pp. 61--71.

\end{thebibliography}

\end{document}